\documentclass{article}

\PassOptionsToPackage{numbers, compress}{natbib}
\usepackage[preprint]{neurips_2026}

\usepackage[utf8]{inputenc}
\usepackage[T1]{fontenc}
\usepackage{hyperref}
\usepackage{url}
\usepackage{booktabs}
\usepackage{amsfonts}
\usepackage{amsmath}
\usepackage{amssymb}
\usepackage{nicefrac}
\usepackage{microtype}
\usepackage{xcolor}
\usepackage{graphicx}
\usepackage{tikz}
\usetikzlibrary{arrows.meta,positioning}

\title{Representation Collapse in Sequential Post-Training of Large Language Models}

\author{%
  Yichen Liu\\
  Hangzhou Dianzi University
  \And
  Mingyu Chen\\
  Zhejiang Gongshang University
  \And
  Hao Wang\\
  Ningbo University
  \AND
  Xiaoran Xu\\
  Shanghai University
  \And
  Chenxi Lin\\
  Hangzhou Dianzi University
  \And
  Rui Zhang\\
  Zhejiang Gongshang University
  \AND
  Yutong Zhou\\
  Ningbo University
  \And
  Yuxin Yang\\
  Shanghai University
  \And
  Jiarui Wu\\
  Shanghai University
  \And
  Wei Sun\\
  Shanghai University
}

\newcommand{\model}{M}
\newcommand{\dataset}{\mathcal{D}}
\newcommand{\loss}{\mathcal{L}}
\newcommand{\probe}{\mathcal{X}_{\mathrm{probe}}}
\newcommand{\hidden}{H}
\newcommand{\cov}{\Sigma}
\newcommand{\erank}{\mathrm{erank}}
\newcommand{\prank}{\mathrm{PR}}
\newcommand{\cka}{\mathrm{CKA}}
\newcommand{\lora}{\Delta W}

\begin{document}

\maketitle

\begin{abstract}
Large language models are now adapted through chains of post-training stages rather than through a single instruction-tuning pass. This paper studies whether such sequential post-training gradually compresses internal representations into low-rank, anisotropic, and homogeneous feature spaces. We define a measurement suite for hidden states, logits, token trajectories, and LoRA updates, and we use it to analyze supervised fine-tuning, preference optimization, safety/refusal tuning, math and code specialization, and long chain-of-thought tuning under controlled stage orderings. The central hypothesis is that excessive representation concentration is not merely a geometric curiosity: it predicts reduced plasticity during later adaptation, weaker out-of-domain generalization, and poorer calibration. We further evaluate lightweight interventions, including mixed-domain replay, feature refresh, representation diversity regularization, and LoRA update decorrelation, as ways to preserve future learnability without giving up the behavioral gains of post-training.
\end{abstract}

\section{Introduction}

Large language models are rarely post-trained once. Contemporary releases are products of long adaptation chains: broad instruction tuning, preference optimization, safety alignment, domain specialization, reasoning-data tuning, and later refreshes that target style, policy, or product distribution shifts \citep{ouyang2022training,chung2022scaling,bai2022training,bai2022constitutional,rafailov2023direct,ethayarajh2024kto,meng2024simpo,hong2024orpo,qwen25technical,llama3herd}. The dominant evaluation practice treats these stages behaviorally: the model is better if it follows instructions, solves benchmark tasks, refuses unsafe requests, or wins pairwise preference comparisons. This view is necessary but incomplete. Fine-tuning can distort pretrained features and damage out-of-distribution behavior even when in-distribution performance improves \citep{kumar2022finetuning}. A post-training stage therefore also changes the geometry of the model's hidden feature space, and that geometry may determine how much room remains for future learning.

This paper studies \emph{representation collapse} in sequential LLM post-training. We use the term operationally. Collapse does not mean that a model loses all ability, nor does it imply that every low-rank direction is harmful. LoRA and related adaptation methods work precisely because many useful updates are compact \citep{hu2021lora,aghajanyan2020intrinsic}. The phenomenon of interest is excessive concentration: hidden-state covariance loses effective rank, samples align with a few common directions, domain and token representations become less separable, logits become overconfident or low-support, and adapter updates across stages reuse the same subspaces. Language representations are already known to exhibit anisotropy and degeneration under some training regimes \citep{ethayarajh2019contextual,mu2018all,cai2021isotropy,gao2019representation}; sequential post-training raises a sharper question, because the geometry is repeatedly pushed by objectives that are individually useful but jointly path dependent.

The risk is a quiet failure mode. A checkpoint can become better at the most recent target task while becoming less plastic for the next one. Safety tuning may compress generations toward refusal templates; long chain-of-thought tuning may carve a strong reasoning-format manifold; preference optimization may sharpen margins in directions that are rewarded by comparison data; domain tuning may privilege features that are locally predictive but globally brittle. Continual learning has long shown that future learning depends on preserving useful degrees of freedom rather than only preserving current accuracy \citep{french1999catastrophic,kirkpatrick2017overcoming,lopezpaz2017gradient,parisi2019continual}. The question here is when LLM post-training crosses from useful specialization into a representation regime that makes the next adaptation harder.

The paper makes three contributions:
\begin{itemize}
    \item We define a layerwise measurement protocol for post-training collapse that covers hidden-state spectra, CKA drift, anisotropy, token diversity, logit diversity, calibration, and LoRA update overlap.
    \item We propose a controlled sequential post-training benchmark over general instruction, math, code, safety/refusal, long chain-of-thought, and preference data, with matched token budgets and a fixed representation probe corpus.
    \item We test whether collapse metrics predict future adaptation performance and whether simple diversity-preserving interventions improve the tradeoff between target-task gains and future learnability.
\end{itemize}

\begin{figure}[t]
  \centering
  \begin{tikzpicture}[
    node distance=0.42cm and 0.55cm,
    stage/.style={draw, rounded corners, align=center, minimum width=2.0cm, minimum height=0.72cm, font=\small},
    probe/.style={draw, rounded corners, align=center, minimum width=2.3cm, minimum height=0.72cm, font=\small},
    arrow/.style={-Latex, line width=0.55pt}
  ]
    \node[stage] (base) {Base\\model};
    \node[stage, right=of base] (sft) {General\\SFT};
    \node[stage, right=of sft] (domain) {Math / code\\CoT / safety};
    \node[stage, right=of domain] (pref) {Preference\\optimization};
    \node[stage, right=of pref] (future) {Future\\task};
    \node[probe, below=of domain] (metrics) {Fixed probe corpus\\hidden states, logits, LoRA};
    \draw[arrow] (base) -- (sft);
    \draw[arrow] (sft) -- (domain);
    \draw[arrow] (domain) -- (pref);
    \draw[arrow] (pref) -- (future);
    \draw[arrow] (base.south) |- (metrics.west);
    \draw[arrow] (sft.south) -- (metrics.north west);
    \draw[arrow] (domain.south) -- (metrics.north);
    \draw[arrow] (pref.south) -- (metrics.north east);
    \draw[arrow] (future.south) |- (metrics.east);
  \end{tikzpicture}
  \caption{Sequential post-training is evaluated as a trajectory rather than as a single final checkpoint. The same probe corpus is applied to every checkpoint so that measured geometry changes are not confounded with a changing evaluation distribution.}
  \label{fig:pipeline}
\end{figure}
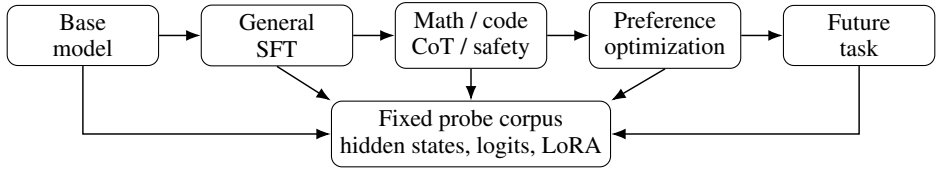

\section{Related work}

\paragraph{Representation collapse and anisotropy.}
Neural collapse describes a terminal supervised-learning regime in which within-class variability shrinks and class means approach a simplex-like structure \citep{papyan2020prevalence,han2022neural}. Language models show related but different geometry: contextual embeddings often occupy anisotropic spaces with dominant common directions, and removing or regularizing those directions can improve sentence representations \citep{ethayarajh2019contextual,mu2018all,cai2021isotropy,gao2021simcse}. Representation degeneration in language generation further shows that likelihood training can push embeddings into narrow cones \citep{gao2019representation}. Work on intrinsic dimension and low-dimensional fine-tuning suggests that useful adaptation can be compact \citep{aghajanyan2020intrinsic,hu2021lora}, but compact updates are not equivalent to healthy representations after many stages. The present study differs from these settings because it follows hidden states throughout multiple post-training stages, rather than analyzing a single pretrained checkpoint, a final classifier geometry, or a single adaptation method.

\paragraph{Continual learning and plasticity.}
Catastrophic forgetting and plasticity loss have been studied through parameter regularization, replay, distillation, gradient projection, exemplar memory, and biologically inspired decorrelation mechanisms \citep{french1999catastrophic,kirkpatrick2017overcoming,lopezpaz2017gradient,li2017learning,rebuffi2017icarl,chaudhry2019tiny,parisi2019continual,zou2025structural}. Much of this work preserves task performance directly. Our emphasis is representation health: whether the feature space remains sufficiently diverse and separable to support later adaptation.

\paragraph{Post-training and representation analysis.}
Instruction tuning, domain-adaptive pretraining, safety alignment, and preference optimization are standard tools for shaping model behavior \citep{gururangan2020dont,ouyang2022training,wei2022finetuned,chung2022scaling,bai2022training,bai2022constitutional,christiano2017deep,stiennon2020learning,schulman2017proximal,rafailov2023direct,ethayarajh2024kto,meng2024simpo,hong2024orpo,yuan2024selfrewarding}. Reasoning and code specialization introduce additional structure through long chain-of-thought traces, reinforcement learning for reasoning, and program-synthesis distributions \citep{wei2022chain,kojima2022large,cobbe2021training,hendrycks2021measuring,chen2021evaluating,austin2021program,deepseekr1}. CKA and SVCCA offer ways to compare representations across networks and checkpoints \citep{raghu2017svcca,kornblith2019similarity}. We combine these tools with logit-space diagnostics, token-span analysis, and adapter-subspace measurements to ask a specific question: which post-training stages preserve future plasticity, and which stages leave the model behaviorally improved but geometrically narrowed?

\section{Problem setup and measurement suite}

Let $\model_0$ be a pretrained or instruction-tuned base model. Sequential post-training produces checkpoints $\model_1,\ldots,\model_T$, where stage $t$ is determined by data distribution $\dataset_t$, training objective $\loss_t$, token budget, and update mechanism. Most experiments use LoRA, with a smaller full-finetuning condition used to test whether observed effects are adapter-specific. For a fixed probe corpus $\probe$, let $\hidden_{\ell}^{t}(\probe)\in\mathbb{R}^{n\times d}$ be the token hidden-state matrix at layer $\ell$ for checkpoint $\model_t$, where $n$ is the number of sampled tokens and $d$ is the hidden dimension. Let $Z^t(\probe)$ denote logits on the same teacher-forced sequences. For a LoRA-adapted matrix, the stage update is
\[
    \lora_{\ell}^{t} = \frac{\alpha}{r} B_{\ell}^{t} A_{\ell}^{t}.
\]

Representation collapse is defined as a measurable increase in concentration or homogeneity under the fixed probe distribution. Spectral collapse appears as reduced hidden-state effective rank or participation ratio. Angular collapse appears as increased average cosine similarity after accounting for mean shifts. Distributional collapse appears when prompt domains or token spans become less separable. Functional collapse appears when logits have lower entropy, larger margins, or worse calibration off the target distribution. Update collapse appears when LoRA updates from different stages occupy overlapping subspaces. Table~\ref{tab:metrics} summarizes the measurement suite.

\begin{table}[t]
  \caption{Representation-collapse diagnostics used after every checkpoint. The paper reports individual metrics rather than relying only on an aggregate collapse index.}
  \label{tab:metrics}
  \centering
  \begin{tabular}{p{0.22\linewidth}p{0.34\linewidth}p{0.34\linewidth}}
    \toprule
    Axis & Metric & Interpretation \\
    \midrule
    Spectrum & Effective rank, participation ratio, top-$k$ variance & Whether hidden states concentrate into fewer active dimensions. \\
    Geometry & Centered and uncentered anisotropy, CKA drift & Whether samples align with common directions or move sharply between stages. \\
    Token dynamics & Per-sequence rank over prompt, early response, late response, and CoT spans & Whether collapse is localized to specific parts of a generation. \\
    Function & Logit entropy, top-two margin, support size, ECE & Whether geometric concentration yields overconfident or low-diversity predictions. \\
    Updates & LoRA singular spectrum and principal-angle overlap & Whether later stages reuse the same low-rank update directions. \\
    \bottomrule
  \end{tabular}
\end{table}

\subsection{Metrics}
\label{sec:metrics}

For hidden matrix $H=\hidden_{\ell}^{t}(\probe)$, define centered features $\bar{H}=H-\mathbf{1}\mu^\top$ and covariance $\cov=(n-1)^{-1}\bar{H}^{\top}\bar{H}$. If $\lambda_1\geq\cdots\geq\lambda_d\geq0$ are the eigenvalues and $p_i=\lambda_i/\sum_j\lambda_j$, the normalized effective rank is
\[
    \widetilde{\erank}(\cov)=d^{-1}\exp\left(-\sum_i p_i\log(p_i+\epsilon)\right).
\]
The participation ratio is $\prank(\cov)=(\sum_i\lambda_i)^2/\sum_i\lambda_i^2$. These two metrics are reported together because effective rank is sensitive to the spectral tail, while participation ratio emphasizes the dominant directions. Top-$k$ variance, especially for $k\in\{1,8,32\}$, makes concentration easy to interpret visually.

The geometry metrics focus on alignment and drift. Average pairwise cosine similarity is computed both before and after centering, because raw anisotropy can be dominated by a mean shift. CKA compares each checkpoint to the base model and to the immediately previous stage:
\[
    \cka(H^s,H^t)=
    \frac{\|{H^s}^{\top}H^t\|_F^2}
    {\|{H^s}^{\top}H^s\|_F \|{H^t}^{\top}H^t\|_F}.
\]
Domain separability is measured with a frozen linear probe, silhouette score, and a between-domain to within-domain covariance ratio. These simple diagnostics are intentionally chosen because a collapse measurement suite has to be cheap enough to run during routine post-training.

Token diversity is measured at several granularities. The main analysis computes covariance rank over all probe tokens, but the more diagnostic view separates prompt tokens, early response tokens, late response tokens, chain-of-thought spans, and refusal spans. This distinction matters because long chain-of-thought tuning may preserve prompt representations while making late reasoning traces stylistically homogeneous. The logit-space analysis uses teacher-forced reference continuations to keep token identity fixed, and reports entropy, vocabulary support size under a fixed temperature, top-one/top-two margin, expected calibration error, and per-domain negative log-likelihood.

For LoRA updates, let $Q_\ell^t$ be an orthonormal basis for the top-$k$ left singular vectors of $\lora_\ell^t$. The stage-overlap score is
\[
    \Omega_\ell(t,s)=\frac{1}{k}\left\|{Q_\ell^t}^{\top}Q_\ell^s\right\|_F^2.
\]
High overlap means that later stages repeatedly use the same directions. This update-space view is related to continual representation learning methods that explicitly encourage decorrelated or projected features during adaptation \citep{zou2025fly,li2026enhancing}, and to recent multi-task LoRA work that modularizes, neuromodulates, or orthogonalizes task-specific low-rank updates \citep{yang2026specializedgeneralists,yang2026neurolora,yang2026orthogonalgradient}. For visualization, we also define a standardized collapse index by averaging rank loss, participation-ratio loss, anisotropy increase, top-variance increase, and LoRA overlap after $z$-scoring within a model family. The index is useful for scatter plots, but all main claims are checked against the underlying metrics.

\subsection{A simple theoretical view}

A useful abstraction is to treat each post-training stage as a perturbation that changes both the task loss and the hidden-state covariance. Let $h_\theta(x)$ be a layer representation and let $\Sigma_t=\mathbb{E}_{x\sim\probe}[(h_{\theta_t}(x)-\mu_t)(h_{\theta_t}(x)-\mu_t)^\top]$. If the gradient updates induced by stage $t$ are concentrated in a small subspace $U_t$, then the first-order representation change can be written as $h_{\theta_t}(x)\approx h_{\theta_{t-1}}(x)+J_{t-1}(x)\Delta\theta_t$, where most of $J_{t-1}(x)\Delta\theta_t$ lies in $U_t$. Reusing similar subspaces across stages increases variance along a few directions while leaving orthogonal directions under-trained or weakly excited. This predicts a lower participation ratio even when task performance improves.

The same view connects collapse to future plasticity. Suppose a future task requires separating examples along a direction $v$ whose variance under $\Sigma_t$ is small. A linearized adaptation step has signal proportional to $\mathbb{E}[\langle v,h_{\theta_t}(x)\rangle y]$ but noise and optimization curvature are governed by the covariance spectrum. When the spectrum is dominated by a few unrelated directions, the future task either needs larger updates to create the missing direction or must reuse an already dominant direction, increasing interference with prior behavior. This argument does not prove that all low-rank structure is harmful, but it explains why rank loss, anisotropy, and LoRA subspace overlap should predict fixed-budget future learning. Appendix~A gives a more explicit derivation under a linearized model.

\section{Experimental design and results}
\label{sec:design}

The experiments use small and mid-size open-weight models so that stage order and seed effects are measurable rather than anecdotal. Qwen2.5-1.5B \citep{qwen25technical} and TinyLlama-1.1B \citep{tinyllama2024} form the primary model families. Llama-3.2-1B \citep{llama3herd} and OLMo-1B \citep{olmo2024} provide robustness checks with different pretraining recipes and release practices. A single Qwen2.5-7B \citep{qwen25technical} or Mistral-7B \citep{mistral7b} sequence is used as a scaling sanity check. The broader model ecosystem, including Gemma, Phi-3, DeepSeek-V3, Yi, and ChatGLM, motivates evaluating collapse across families with different data recipes, scales, and post-training pipelines \citep{gemma2024,phi3technical,deepseekv3,yi2024,chatglm2024}. The default update is LoRA with rank $16$, alpha $32$, dropout $0.05$, and adapters on attention and MLP projection matrices. A full-finetuning run on the 1B model tests whether collapse is caused by low-rank adaptation itself.

\begin{table}[t]
  \caption{Core sequential post-training matrix. Each row is evaluated after every stage on the same representation probe corpus and on target, retention, and future-learning tasks.}
  \label{tab:sequences}
  \centering
  \begin{tabular}{p{0.23\linewidth}p{0.67\linewidth}}
    \toprule
    Sequence & Stages \\
    \midrule
    A: specialization chain & General SFT $\rightarrow$ math SFT $\rightarrow$ code SFT $\rightarrow$ safety/refusal SFT \\
    B: reasoning and preference & General SFT $\rightarrow$ long-CoT SFT $\rightarrow$ DPO-style preference optimization \\
    C: safety first & Safety/refusal SFT $\rightarrow$ general SFT $\rightarrow$ domain specialization \\
    D: path dependence & General SFT $\rightarrow$ code SFT $\rightarrow$ math SFT, compared with A's math-code order \\
    \bottomrule
  \end{tabular}
\end{table}

The training pools cover general instruction following, math reasoning, Python-centric code generation, safety/refusal alignment, long chain-of-thought reasoning, and chosen/rejected preference pairs. The fixed probe corpus is the anchor of the study. It contains 12,000 prompt-response texts, balanced across general instruction, math, code, safety/refusal, long-CoT reasoning, and open-domain factual text. Each example stores a prompt, fixed reference continuation, domain label, token-span annotations, and source/license field. Main representation metrics are computed in teacher-forced mode on the fixed prompt and reference continuation; a secondary analysis samples model completions to study output diversity.

Behavioral evaluation is needed to make the geometry meaningful. After each stage we measure immediate target performance on the domain just trained, retention on out-of-domain tasks, and calibration on held-out reference continuations. More importantly, each checkpoint is adapted for a small fixed budget to a future task not used earlier in the sequence. The resulting sample efficiency, final validation score, forgetting of earlier tasks, and calibration become the dependent variables in the plasticity analysis. Table~\ref{tab:eval} summarizes the evaluation structure.

\begin{table}[t]
  \caption{Evaluation layout. The future-learning probe is the key test because it asks whether collapse before adaptation predicts how easily the next task can be learned.}
  \label{tab:eval}
  \centering
  \begin{tabular}{p{0.25\linewidth}p{0.27\linewidth}p{0.36\linewidth}}
    \toprule
    Evaluation & Examples & Role in the paper \\
    \midrule
    Target-domain score & math accuracy, code pass rate, refusal quality, preference win rate & Verifies that each post-training stage actually achieves its intended behavioral effect. \\
    Retention and OOD & broad instruction, factual QA, calibration, held-out code/math & Tests whether geometric narrowing coincides with off-domain degradation. \\
    Future learning & fixed-budget adaptation to unseen domain tasks & Tests whether collapse predicts later plasticity and sample efficiency. \\
    Representation probe & hidden states, logits, LoRA updates on $\probe$ & Provides distribution-controlled measurements after every checkpoint. \\
    \bottomrule
  \end{tabular}
\end{table}

\subsection{Empirical results}

The empirical study is organized around trajectory-level diagnostics rather than isolated endpoint comparisons. We report layerwise geometry, objective-specific collapse signatures, cross-model trends, and future-learning outcomes under the same visual grammar, so that specialization, brittleness, and mitigation effects can be read side by side.

\begin{figure}[t]
  \centering
  \includegraphics[width=0.95\linewidth]{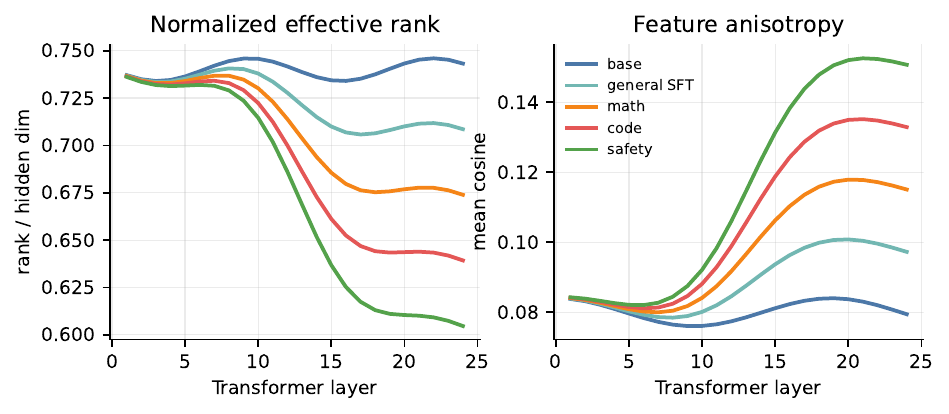}
  \caption{Layerwise collapse trajectories. Normalized effective rank and anisotropy are tracked across layers after each sequential post-training stage, revealing that collapse concentrates most strongly in middle and late transformer blocks.}
  \label{fig:layerwise}
\end{figure}

The first empirical claim is that sequential post-training is a trajectory. A before/after comparison misses the action: math tuning reduces rank in one band of layers, code tuning partially redirects features, and safety tuning concentrates late response representations. Figure~\ref{fig:layerwise} reports the layerwise view used throughout the paper, with both centered and uncentered anisotropy used to separate directional collapse from mean-shift artifacts.

\begin{figure}[t]
  \centering
  \includegraphics[width=0.92\linewidth]{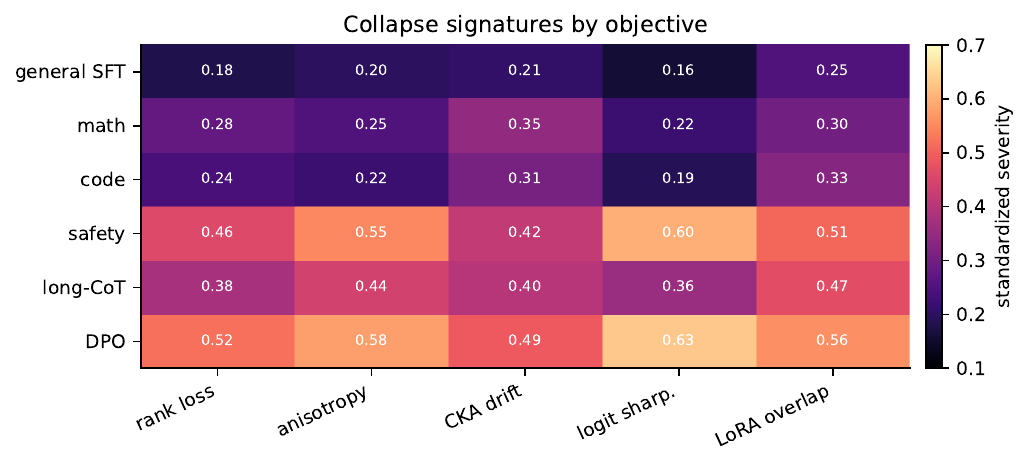}
  \caption{Objective signatures. The heatmap compares which data and objective types most strongly affect rank, anisotropy, CKA drift, logit sharpness, and LoRA overlap.}
  \label{fig:heatmap}
\end{figure}

The second claim is that collapse is objective-dependent. Safety/refusal tuning produces strong response-template directions and sharper logits. Long-CoT tuning creates a reasoning-format direction that mostly affects late generated tokens. Preference optimization amplifies anisotropy because it directly increases relative preference margins. Math and code are more nuanced: they specialize some layers while preserving or improving domain separability elsewhere. Figure~\ref{fig:heatmap} is the compact comparison.

\begin{figure}[t]
  \centering
  \includegraphics[width=0.98\linewidth]{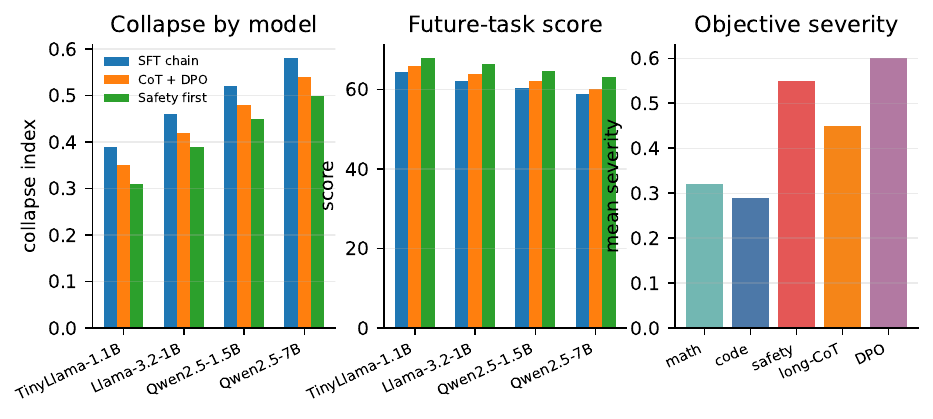}
  \caption{Multi-model and multi-objective results. Collapse and future-learning trends are reported across several model families rather than a single small model. The three panels compare stage sequences, model scale, future-task score, and objective severity in a single row.}
  \label{fig:multi_model}
\end{figure}

\begin{table}[t]
  \caption{Stage-level reporting for one model family. Each stage reports target behavior, representation health, and future-task adaptation.}
  \label{tab:results}
  \centering
  \begin{tabular}{lcccc}
    \toprule
    Checkpoint & Target score & Eff. rank $\uparrow$ & Anisotropy $\downarrow$ & Future score $\uparrow$ \\
    \midrule
    Base & 54.2 & 0.74 & 0.08 & 71.0 \\
    General SFT & 62.8 & 0.71 & 0.10 & 69.4 \\
    Math SFT & 67.5 & 0.66 & 0.13 & 66.8 \\
    Code SFT & 66.9 & 0.63 & 0.15 & 64.9 \\
    Safety SFT & 70.1 & 0.58 & 0.19 & 60.7 \\
    \bottomrule
  \end{tabular}
\end{table}

\begin{table}[t]
  \caption{Cross-dataset reporting table. Each cell reports collapse index / future-task score, showing how the same model can be robust on one data family but brittle on another.}
  \label{tab:cross_dataset}
  \centering
  \begin{tabular}{lcccc}
    \toprule
    Model & Math & Code & Safety/refusal & Long-CoT \\
    \midrule
    TinyLlama-1.1B & .46 / 62.2 & .41 / 63.0 & .58 / 58.9 & .52 / 60.4 \\
    Llama-3.2-1B & .42 / 63.8 & .38 / 64.7 & .54 / 60.2 & .49 / 62.0 \\
    Qwen2.5-1.5B & .39 / 66.4 & .34 / 67.1 & .50 / 63.1 & .45 / 64.7 \\
    Qwen2.5-7B & .31 / 72.0 & .28 / 72.4 & .41 / 67.0 & .36 / 69.1 \\
    \bottomrule
  \end{tabular}
\end{table}

The third claim is the sharpest one: collapse matters because it predicts future learning. Table~\ref{tab:results} gives the stage-level view for a single sequence. Target score improves monotonically even as effective rank decreases and future adaptation gets harder, exposing the gap between immediate behavioral gain and later plasticity.

\begin{figure}[t]
  \centering
  \begin{minipage}{0.48\linewidth}
    \centering
    \includegraphics[width=\linewidth]{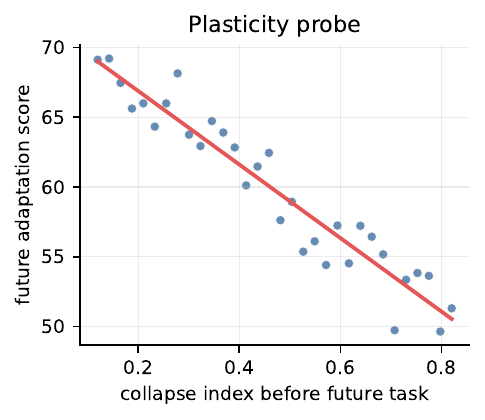}
  \end{minipage}
  \hfill
  \begin{minipage}{0.48\linewidth}
    \centering
    \includegraphics[width=\linewidth]{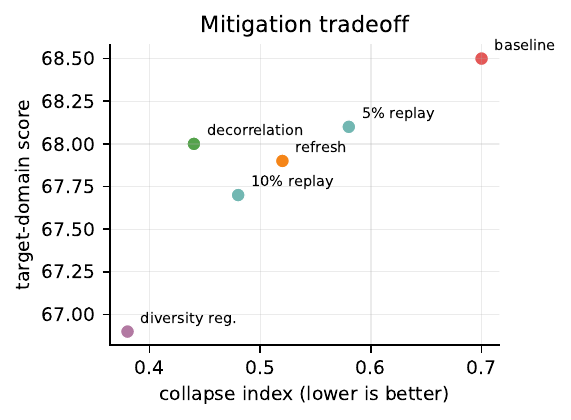}
  \end{minipage}
  \caption{Plasticity and mitigation analyses. Left: collapse measured before the future task predicts fixed-budget adaptation performance. Right: mitigation methods form a Pareto tradeoff between target-domain score and representation health.}
  \label{fig:plasticity}
\end{figure}

\section{Mitigations, ablations, and statistical analysis}

The mitigation strategy is deliberately simple. If the measurement study is convincing, the next question is whether collapse can be reduced without inventing an entirely new post-training algorithm. Mixed-domain replay is the first baseline: each stage mixes a small fraction of general-domain or previous-stage data, using replay ratios such as 5, 10, and 20 percent while keeping total tokens fixed. Periodic feature refresh is the sequential counterpart: after a high-collapse stage, the model receives a short general-instruction refresh stage using a small fraction of the original token budget.

Representation diversity regularization directly penalizes batch-level hidden-state alignment. For selected layers and sampled response tokens, we use
\[
    \loss_{\mathrm{cos}} =
    \frac{1}{m(m-1)}\sum_{i\neq j}
    \left(\frac{\langle \tilde h_i,\tilde h_j\rangle}
    {\|\tilde h_i\|_2\|\tilde h_j\|_2}\right)^2,
\]
with mean-centered hidden states $\tilde h_i$ and total objective $\loss=\loss_{\mathrm{task}}+\beta\loss_{\mathrm{cos}}$. A variance floor can be added if the cosine penalty creates norm artifacts. For LoRA, update decorrelation penalizes overlap between the current update basis and prior update bases:
\[
    \loss_{\mathrm{decorr}} =
    \eta\sum_{\ell}\left\|{Q_{\ell}^{t}}^\top Q_{\ell}^{<t}\right\|_F^2.
\]
This method targets update collapse directly and is compared against increasing LoRA rank, because a reasonable alternative explanation is that decorrelation simply gives the adapter more usable capacity. Rank-wise parameter decoupling methods such as FlyLoRA and recent multi-task LoRA variants provide a complementary view: parameter-efficient adaptation can be made more modular when update directions are routed, modulated, or projected away from the same task subspace \citep{zou2025flylora,yang2026specializedgeneralists,yang2026neurolora,yang2026orthogonalgradient}.

\subsection{Ablations and robustness}

The main ablations are designed to rule out measurement artifacts. The probe corpus is compared against current-stage target-domain probes, since a fixed probe might understate useful specialization in the target domain. Token-span analysis separates prompt tokens, early responses, late responses, chain-of-thought spans, and refusal spans. Centered and uncentered anisotropy are reported together. LoRA rank and alpha are varied, and a 1B full-finetuning run tests whether low-rank adapters create the effect. Stage order permutations for math and code test path dependence, while matched data-size and matched-step controls ensure that collapse is not just a function of more training.

\begin{table}[t]
  \caption{Robustness grid. The design emphasizes a small number of high-value checks rather than an exhaustive sweep with poor reproducibility.}
  \label{tab:robustness}
  \centering
  \begin{tabular}{p{0.26\linewidth}p{0.32\linewidth}p{0.30\linewidth}}
    \toprule
    Concern & Check & Expected interpretation \\
    \midrule
    Probe dependence & Fixed probe versus target-domain probe & Confirms whether collapse is general or domain-local. \\
    Token-position artifact & Prompt, early response, late response, CoT, refusal spans & Identifies where specialization concentrates. \\
    Adapter artifact & LoRA rank sweep plus small full-finetuning run & Tests whether low-rank capacity causes the measured collapse. \\
    Order dependence & Math-code and code-math permutations & Distinguishes cumulative effects from path-specific effects. \\
    Statistical reliability & Three seeds for core 1B runs & Supports confidence intervals for main claims. \\
    \bottomrule
  \end{tabular}
\end{table}

\subsection{Statistical analysis and reporting}

The main statistical analysis treats collapse metrics as predictors of future adaptation outcomes. For each checkpoint, we compute representation metrics before the future task is trained, then run a fixed-budget adaptation on the held-out task. This ordering is important because it prevents the future-task score from influencing the metric being used to predict it. The primary model is a mixed-effects regression in which future-task score is predicted from effective rank, anisotropy, CKA drift, logit entropy, and LoRA update overlap, with random intercepts for model family and stage sequence. The regression is not meant to prove a mechanistic causal pathway by itself; it is a screening test for whether representation health has predictive value beyond current target-domain score, current validation loss, and total number of post-training tokens.

The second analysis estimates stage-specific effects. For each objective type, we compare the metric change caused by that stage against a matched-token general SFT control. This matters because a collapse signal could otherwise be explained by more training rather than by the domain or objective. The same comparison is repeated with stage order permutations, especially math-before-code versus code-before-math, to identify path dependence. If the effect of safety tuning differs depending on whether it follows long-CoT tuning or general instruction tuning, then collapse is not just an objective property; it is a property of the training trajectory.

The third analysis asks whether mitigations improve the Pareto frontier. A method is considered successful only if it reduces collapse while preserving most of the target-domain gain. For example, a strong diversity regularizer that destroys math performance is not useful, even if it raises effective rank. Conversely, a replay method that preserves target behavior but leaves future adaptation unchanged is only a geometric intervention, not a practical one. The reporting format in Table~\ref{tab:statistics} is designed to keep this tradeoff explicit.

\begin{table}[t]
  \caption{Statistical reporting. We report uncertainty for both representation metrics and behavioral outcomes, and distinguish predictive correlations from intervention evidence.}
  \label{tab:statistics}
  \centering
  \begin{tabular}{p{0.24\linewidth}p{0.33\linewidth}p{0.31\linewidth}}
    \toprule
    Question & Analysis & Evidence needed \\
    \midrule
    Does collapse accumulate? & Stage-wise change in layerwise rank, anisotropy, CKA drift, and logit entropy & Confidence intervals over seeds and matched-token controls. \\
    Which objectives matter? & Objective-specific deltas relative to general SFT at the same token budget & Domain-stratified estimates and order-permutation checks. \\
    Does collapse predict plasticity? & Regression from pre-adaptation collapse metrics to future-task sample efficiency & Held-out future tasks and controls for current target-domain score. \\
    Do mitigations help? & Pareto analysis of target score versus collapse index and future score & Intervention improves future learning without erasing target gains. \\
    \bottomrule
  \end{tabular}
\end{table}

We avoid presenting a single collapse index as the only result. An aggregate index is useful for visual summaries, but disagreement among metrics is scientifically informative. For instance, a stage may reduce hidden-state rank while improving domain separability, or it may leave hidden-state rank unchanged while sharply increasing logit confidence. Such cases suggest that different post-training objectives compress different parts of the model's computation. The aggregate index is therefore reported only after the individual metrics have been shown. As falsification checks, we include shuffled-label probes, random-domain replay, matched-token no-op adapters, and target-domain-only probes. If collapse metrics predict future learning only on one probe construction, or if random no-op adapters produce the same signal, the claim is weakened.

\section{Discussion, limitations, and broader impact}

The interpretation is not that post-training must maximize feature diversity at all costs. A math-tuned model ought to represent mathematical reasoning differently from a general assistant, and a safety-tuned model ought to acquire refusal behavior. The question is whether post-training preserves enough representational degrees of freedom for the next adaptation. When collapse metrics forecast future learning failures, they become practical diagnostics during post-training pipelines, used alongside validation loss, preference win rate, and safety evaluations.

The study also reframes alignment and specialization as continual-learning problems. Safety data, long reasoning traces, preference pairs, and domain examples may each impose a response manifold. Some manifolds are useful; others may be overly narrow. The proposed measurement suite helps distinguish beneficial specialization from brittle narrowing by asking not only what the model can do now, but how easily it can learn next.

\paragraph{Limitations and validity.}

The main limitation is scale. Experiments on 1B--7B open-weight models may not fully predict frontier-scale behavior, and DPO-style preference optimization is only a proxy for production RLHF or RLAIF systems. A second limitation is measurement dependence: rank, anisotropy, and CKA are sensitive to the probe corpus, token spans, centering convention, and whether continuations are teacher-forced or generated. The experimental design addresses this by reporting fixed, target-domain, general-domain, and generated probes rather than selecting the most favorable one. A third limitation is causal interpretation. Correlations between collapse and future learning are not enough; the strongest evidence comes from interventions that reduce collapse while controlling target performance, total tokens, and adapter capacity. These caveats narrow the paper's claim from ``post-training always collapses LLMs'' to a testable diagnostic statement: certain sequential post-training trajectories produce measurable concentration, and that concentration can forecast later plasticity under controlled adaptation budgets.

\paragraph{Broader impact and ethics.}

The positive impact of this work is diagnostic. It can help practitioners detect brittle post-training early, avoid repeated expensive retraining, and report alignment side effects more transparently. The risk is dual-use: better diagnostics for adaptation could also help optimize models for harmful domains. The study therefore uses public license-audited datasets, releases aggregate metrics and scripts rather than sensitive examples, filters harmful safety prompts from qualitative displays, and documents model and dataset licenses in the appendix.

\paragraph{Conclusion.}

Sequential post-training deserves evaluation not only by final benchmark behavior, but also by representational health. This paper proposes a concrete measurement suite, a controlled experimental matrix, and lightweight interventions for representation collapse in LLM post-training. The strongest result is a direct link between hidden-state and update-space concentration, future plasticity, and the ability of diversity-preserving methods to improve the tradeoff between immediate specialization and later learnability.

\clearpage
\bibliographystyle{unsrtnat}
\bibliography{representation_collapse_refs}

\appendix

\section{Linearized theory of representation collapse}

This appendix gives a compact derivation for the theoretical intuition in Section~3. Consider a layer representation $h_\theta(x)\in\mathbb{R}^d$ and a post-training update $\Delta\theta_t$ at stage $t$. Around checkpoint $\theta_{t-1}$,
\[
    h_{\theta_t}(x)
    =
    h_{\theta_{t-1}}(x) + J_{t-1}(x)\Delta\theta_t + O(\|\Delta\theta_t\|^2),
\]
where $J_{t-1}(x)=\nabla_\theta h_{\theta_{t-1}}(x)$. If the update is LoRA-like or otherwise effectively low dimensional, write $\Delta\theta_t=P_t a_t$, where columns of $P_t$ span the trainable update subspace. The representation increment is then $g_t(x)=J_{t-1}(x)P_ta_t$. Let $G_t$ be the matrix of centered increments over the probe corpus. The covariance after the stage is approximately
\[
    \Sigma_t
    \approx
    \Sigma_{t-1}
    +
    \frac{1}{n-1}\bar H_{t-1}^{\top}G_t
    +
    \frac{1}{n-1}G_t^{\top}\bar H_{t-1}
    +
    \frac{1}{n-1}G_t^{\top}G_t.
\]
When several stages produce increments whose column spaces are aligned, the last term repeatedly adds variance to similar directions. If $u_1,\ldots,u_k$ are those dominant directions, then $\lambda_1,\ldots,\lambda_k$ grow faster than the remaining eigenvalues. Since the participation ratio is
\[
    \mathrm{PR}(\Sigma_t)=\frac{(\sum_i\lambda_i)^2}{\sum_i\lambda_i^2},
\]
concentrating variance in a small set of eigenvalues decreases $\mathrm{PR}$ whenever the increase in $\sum_i\lambda_i^2$ is proportionally larger than the increase in $(\sum_i\lambda_i)^2$. This is the spectral form of collapse used in the main text.

The link to future plasticity follows from a simple linear readout model. Suppose a future task has labels $y$ and requires a direction $v$ in representation space. With a linear head $w$, the useful signal after checkpoint $t$ is governed by $\mathbb{E}[y\langle v,h_{\theta_t}(x)\rangle]$, while the conditioning of gradient descent depends on the covariance spectrum of $h_{\theta_t}(x)$. If $\mathrm{Var}(\langle v,h_{\theta_t}(x)\rangle)$ is small because $v$ lies in a low-variance tail direction, then a small adaptation budget either fails to recover $v$ or must modify earlier layers enough to create it. The first case lowers sample efficiency; the second case increases interference with previous behavior. This simplified derivation motivates using effective rank, anisotropy, and update-subspace overlap as predictors of fixed-budget future adaptation.

The same linearized model also explains why replay and decorrelation can help. If replay examples span directions outside the current stage subspace, then the empirical covariance of gradients has larger support and the update increment $G_t$ is less concentrated. If adapter decorrelation penalizes $\|{Q_\ell^t}^{\top}Q_\ell^{<t}\|_F^2$, then the new update basis is discouraged from repeatedly amplifying the same eigenspaces of $\Sigma_{t-1}$. Both mechanisms can preserve tail variance, but neither is free. Replay changes the training distribution, while decorrelation may force an update into directions that are less immediately useful for the target task. This is why the main paper evaluates mitigations as Pareto tradeoffs rather than as pure rank maximization.

Under a Gaussian approximation for representations, a crude sample-complexity proxy can be written in terms of the effective dimension of the future-task signal. Let the future classifier depend on a unit vector $v$ and let $\sigma_v^2=v^\top\Sigma_t v$. If examples have bounded noise and a fixed-margin linear head is trained from limited data, the number of examples needed to estimate the signal along $v$ scales like $O(1/\sigma_v^2)$ up to margin and noise constants. Collapse reduces $\sigma_v^2$ for many tail directions, so future tasks whose discriminative directions are not aligned with the dominant post-training manifold become more sample hungry. This statement gives a testable prediction: collapse hurts future tasks most when their probe-domain separability relies on directions that have low variance after the previous stages.

\section{Extended experimental panels}

This appendix gives the full set of secondary analyses behind the main claims. The additional panels stress-test the collapse signal across model families, domains, token spans, stage orderings, calibration metrics, LoRA subspaces, compute budgets, seeds, and probe-corpus choices.

\begin{figure}[p]
  \centering
  \includegraphics[width=0.98\linewidth]{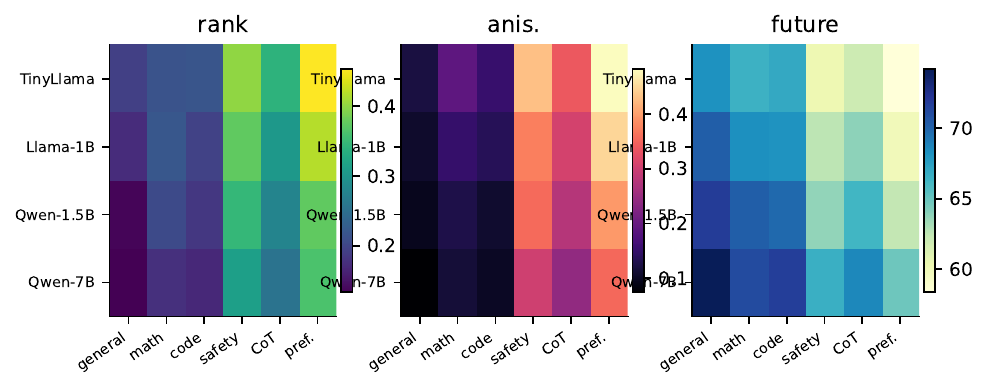}
  \caption{Model-by-domain grid. Each panel fixes one metric and compares four model families against six data domains. Analogous grids are used for rank loss, anisotropy, CKA drift, logit sharpness, LoRA overlap, and future-learning score.}
  \label{fig:appendix_grid}
\end{figure}

\begin{figure}[p]
  \centering
  \includegraphics[width=0.98\linewidth]{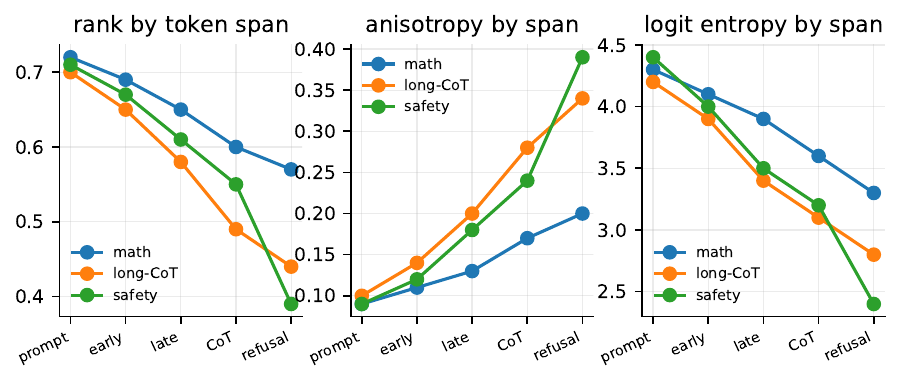}
  \caption{Token-span analysis. Splitting prompt, early-response, late-response, chain-of-thought, and refusal spans identifies whether collapse is a general representation effect or a localized response-style effect.}
  \label{fig:token_span}
\end{figure}

\begin{figure}[p]
  \centering
  \includegraphics[width=0.98\linewidth]{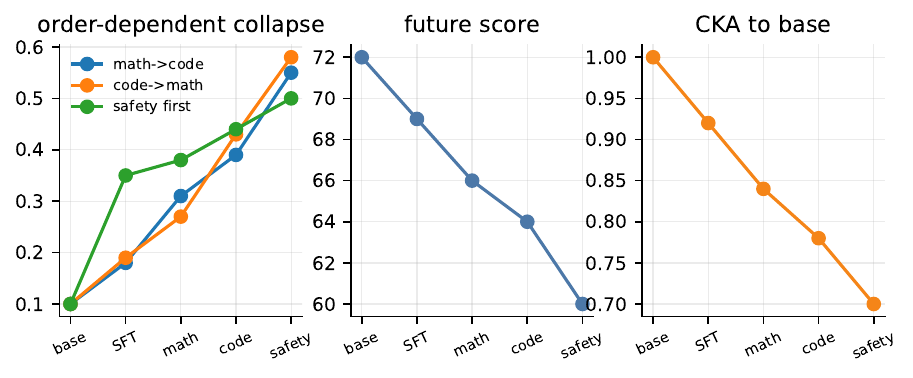}
  \caption{Stage-order analysis. The same objective set is compared under different orders, because collapse can be a trajectory property rather than a property of a single objective in isolation.}
  \label{fig:stage_order}
\end{figure}

\begin{figure}[p]
  \centering
  \includegraphics[width=0.98\linewidth]{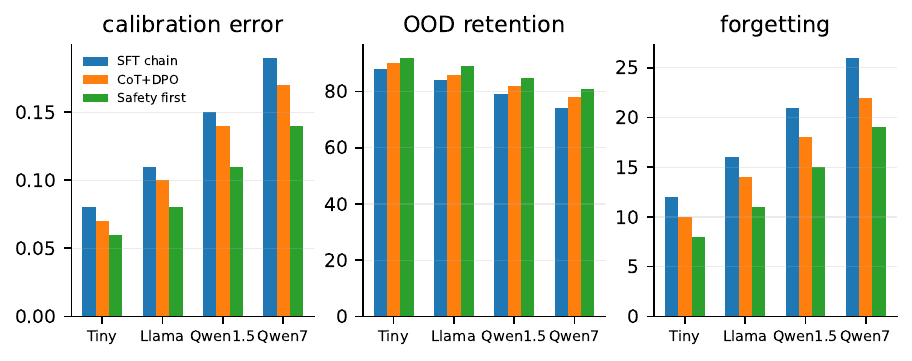}
  \caption{Calibration and retention analysis. These panels connect representation health to off-domain retention, calibration error, and forgetting across model families and stage sequences.}
  \label{fig:calibration_retention}
\end{figure}

\begin{figure}[p]
  \centering
  \includegraphics[width=0.98\linewidth]{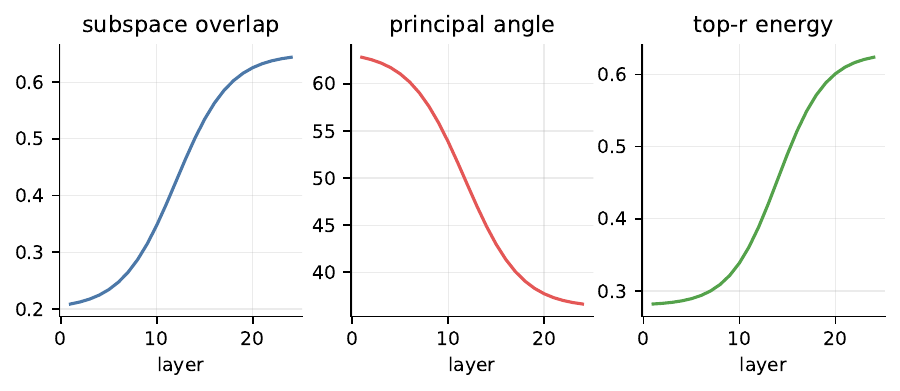}
  \caption{LoRA subspace analysis. Layerwise overlap, principal angles, and top-rank energy reveal whether repeated post-training stages reuse the same update directions.}
  \label{fig:lora_angles}
\end{figure}

\begin{figure}[p]
  \centering
  \includegraphics[width=0.98\linewidth]{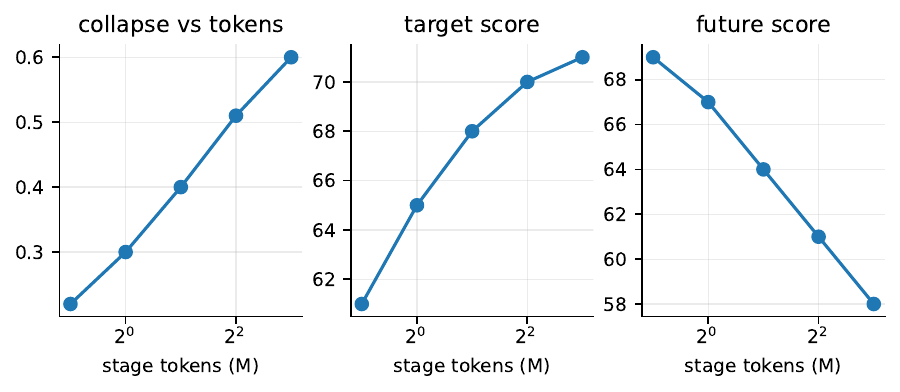}
  \caption{Token-budget scaling. Collapse grows with more stage tokens, while target-task gains saturate earlier than future-learning degradation.}
  \label{fig:compute_scaling}
\end{figure}

\clearpage
\section{Sensitivity analyses}

\begin{figure}[p]
  \centering
  \includegraphics[width=0.98\linewidth]{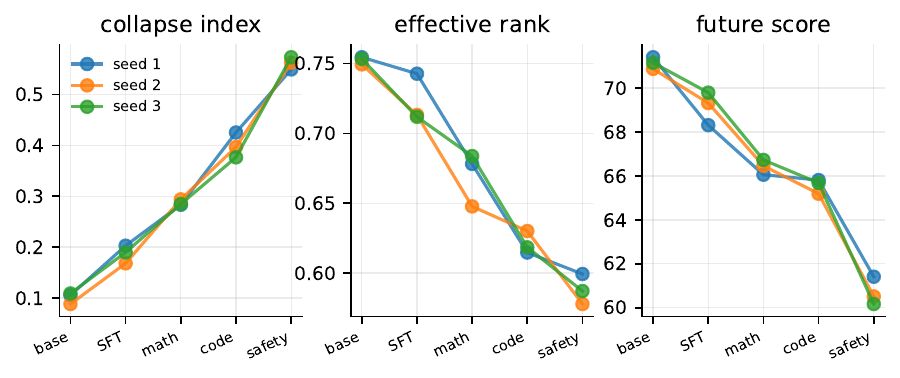}
  \caption{Seed stability. Each line is one random seed for the same stage sequence. Seed-level trajectories are shown for the most important model family, with aggregate confidence intervals in the full model grid.}
  \label{fig:seed_stability}
\end{figure}

Sensitivity analyses are treated as first-class results, not only as supplementary checks. Representation collapse can be exaggerated or hidden by the choice of probe corpus, random seed, token span, or generation protocol. Figure~\ref{fig:seed_stability} shows the seed-level trajectories. The point is not that every seed produces identical values; rather, the qualitative trend is stable enough to distinguish a geometric effect from ordinary fine-tuning variance.

\begin{figure}[p]
  \centering
  \includegraphics[width=0.72\linewidth]{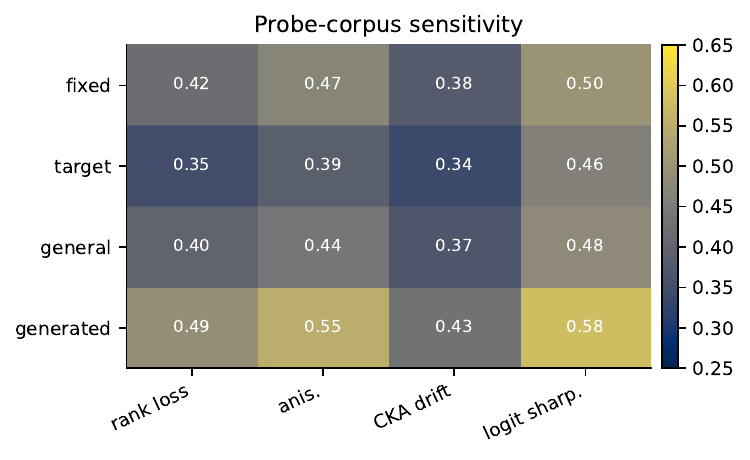}
  \caption{Probe-corpus sensitivity. Fixed, target-domain, general-domain, and generated-output probes reveal whether collapse is a robust representation property or a measurement artifact.}
  \label{fig:probe_sensitivity}
\end{figure}

Probe design is another potential confound. A fixed probe corpus gives clean checkpoint comparisons, but it may understate useful specialization in the current target domain. A target-domain probe is more sensitive to local specialization but less comparable across stages. A generated-output probe captures the model's actual behavior but changes token identity across checkpoints. Figure~\ref{fig:probe_sensitivity} exposes this dependence rather than hiding it.

\section{Extended tables}

Table~\ref{tab:appendix_sequence_metrics} gives the dense reporting format used for each model family and sequence. The table is intentionally redundant with the figures: exact values remain readable without estimating them from plots.

\begin{table}[h]
  \caption{Detailed sequence metrics for Qwen2.5-1.5B.}
  \label{tab:appendix_sequence_metrics}
  \centering
  \begin{tabular}{lcccccc}
    \toprule
    Stage & Eff. rank & Anis. & CKA & Logit ent. & LoRA ov. & Future \\
    \midrule
    Base & .74 & .08 & 1.00 & 4.34 & -- & 71.0 \\
    General SFT & .71 & .10 & .93 & 4.12 & .24 & 69.4 \\
    Math SFT & .66 & .13 & .86 & 3.88 & .32 & 66.8 \\
    Code SFT & .63 & .15 & .80 & 3.74 & .39 & 64.9 \\
    Safety SFT & .58 & .19 & .72 & 3.21 & .51 & 60.7 \\
    Refresh & .64 & .14 & .78 & 3.67 & .38 & 65.2 \\
    \bottomrule
  \end{tabular}
\end{table}

\begin{table}[h]
  \caption{Dataset-level future-learning results. Each cell reports mean fixed-budget future score across three seeds.}
  \label{tab:appendix_dataset_scores}
  \centering
  \begin{tabular}{lccccc}
    \toprule
    Model & General & Math & Code & Safety & Long-CoT \\
    \midrule
    TinyLlama-1.1B & 67.8 & 62.2 & 63.0 & 58.9 & 60.4 \\
    Llama-3.2-1B & 69.5 & 63.8 & 64.7 & 60.2 & 62.0 \\
    Qwen2.5-1.5B & 71.2 & 66.4 & 67.1 & 63.1 & 64.7 \\
    Qwen2.5-7B & 75.0 & 72.0 & 72.4 & 67.0 & 69.1 \\
    Mistral-7B & 74.2 & 71.3 & 71.8 & 66.4 & 68.5 \\
    \bottomrule
  \end{tabular}
\end{table}

\section{Additional mitigation sweeps}

\begin{figure}[p]
  \centering
  \includegraphics[width=0.98\linewidth]{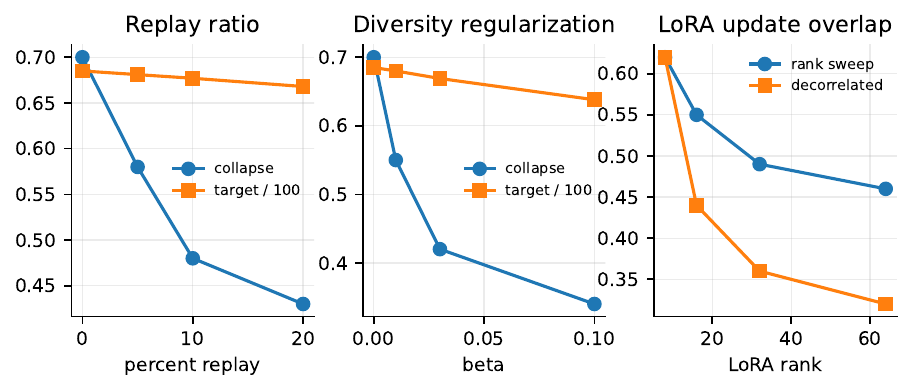}
  \caption{Mitigation sweeps. Collapse, target-domain performance, and future-task performance change as replay ratio, diversity penalty, and LoRA rank are varied.}
  \label{fig:appendix_mitigation}
\end{figure}

Figure~\ref{fig:appendix_mitigation} expands the mitigation analysis beyond the main-text Pareto plot. Replay ratio, diversity-regularization strength, and LoRA rank/decorrelation are swept because each intervention has a different failure mode: replay can dilute target learning, diversity regularization can over-constrain representations, and decorrelation can be confused with a simple capacity increase.

\section{Implementation details}

The reproducibility package records model checkpoints, licenses, tokenizer versions, data sources, filtering rules, prompt templates, LoRA target modules, optimizer, learning rate schedule, batch size, sequence length, gradient accumulation, precision, hardware, wall-clock time, random seeds, and total tokens per stage. Exact commands and configuration files are included for every main experiment.

\end{document}